\documentclass[runningheads]{llncs}
\usepackage[utf8]{inputenc} 
\usepackage{booktabs}
\usepackage[T1]{fontenc}
\usepackage{tabularx} 
\usepackage{graphicx} 
\usepackage{amsmath}
\usepackage{siunitx} 
\usepackage{colortbl}
\usepackage[table]{xcolor}

\definecolor{g1}{HTML}{D6D5C1} 
\definecolor{g2}{HTML}{FAFAF6}
\definecolor{g3}{HTML}{ECEBE2}

\usepackage{graphicx,verbatim}
\usepackage{multirow}

\begin{document}
\title{Pancreas Part Segmentation under Federated Learning Paradigm}
%

 %% Added for anonymized MICCAI 2025 submission
\author{
Ziliang Hong\inst{1} \and
Halil Ertugrul Aktas\inst{1} \and
Andrea Mia Bejar\inst{1} \and
Katherine Wu\inst{7}\and
Hongyi Pan\inst{1} \and
Gorkem Durak\inst{1} \and
Zheyuan Zhang\inst{1} \and
Sait Kayali\inst{8} \and
Temel Tirkes\inst{2} \and
Federica Proietto Salanitri\inst{3} \and
Concetto Spampinato\inst{3} \and
Michael Goggins\inst{4} \and
Tamas Gonda\inst{5} \and
Candice Bolan\inst{6} \and
Raj Keswani\inst{1} \and
Frank Miller\inst{1} \and
Michael Wallace\inst{6} \and
Ulas Bagci\inst{1}
}

\authorrunning{Z. Hong et al.}
\institute{Northwestern University, Evanston, IL, USA\\
% \email{\{zilianghong2029, halilertugrul.aktas, andrea.bejar, hongyi.pan, gorkem.durak, zheyuan.zhang, raj-keswani, fmiller, ulas.bagci\}@northwestern.edu}
\and
Indiana University–Purdue University Indianapolis, Indianapolis, IN, USA\\
% \email{atirkes@iupui.edu}
\and
University of Catania, Catania, Italy\\
% \email{\{federica.proiettosalanitri, concetto.spampinato\}@unict.it}
\and
Johns Hopkins University, Baltimore, MD, USA\\
% \email{mgoggins@jhmi.edu}
\and
New York University Langone Health, NY, USA\\
% \email{Tamas.Gonda@nyulangone.org}
\and
Mayo Clinic, Rochester, MN, USA\\
% \email{\{Bolan.Candice, wallace.michael\}@mayo.edu}
\and
University of Illinois Chicago, Chicago, IL, USA\\
\and
Istanbul University, Istanbul, Turkey\\
}
    
\maketitle % typeset the header of the contribution

\begin{abstract}

We present the first federated learning (FL) approach for pancreas part (\textit{head, body, tail}) segmentation in MRI, addressing a critical clinical challenge as a significant innovation. Pancreatic diseases exhibit marked regional heterogeneity—cancers predominantly occur in the \textit{head} region while chronic pancreatitis causes tissue loss in the \textit{tail}—making accurate segmentation of the organ into head, body, and tail regions essential for precise diagnosis and treatment planning. This segmentation task remains exceptionally challenging in MRI due to variable morphology, poor soft-tissue contrast, and anatomical variations across patients. Our novel contribution tackles two fundamental challenges: first, the technical complexity of pancreas part delineation in MRI, and second the data scarcity problem that has hindered prior approaches. We introduce a privacy-preserving FL framework that enables collaborative model training across seven medical institutions without direct data sharing, leveraging a diverse dataset of 711 T1W and 726 T2W MRI scans. Our key innovations include: (1) a systematic evaluation of three state-of-the-art segmentation architectures (U-Net, Attention U-Net,Swin UNETR) paired with two FL algorithms (FedAvg, FedProx), revealing Attention U-Net with FedAvg as optimal for pancreatic heterogeneity, which was never been done before; (2) a novel anatomically-informed loss function prioritizing region-specific texture contrasts in MRI. Comprehensive evaluation demonstrates that our approach achieves clinically viable performance despite training on distributed, heterogeneous datasets.

% Authors must provide keywords and are not allowed to remove this Keyword section.

\end{abstract}
\section{Introduction}

Pancreatic diseases represent significant global health challenges, with their impact magnified by the rising prevalence of obesity and metabolic syndrome~\cite{rawla2019epidemiology,eibl2018diabetes}. A critical yet underexplored aspect of these pathologies is their marked regional heterogeneity across the pancreas head, body, and tail—a characteristic that fundamentally alters disease presentation, progression, and treatment approaches~\cite{schoennagel2011diffusion}.

Recent histological investigations have revealed that the pancreatic tail contains a higher concentration of insulin-producing cells with enhanced endocrine function compared to the head and body regions~\cite{wang2013regional,ravi2021redefining}. This anatomical specialization creates distinct disease patterns: pancreatic cancers predominantly develop in the head region~\cite{australian2018defining}, while chronic pancreatitis often causes severe tissue loss and fibrosis in the tail~\cite{desouza2019pancreas}. Similarly, pancreatic cystic lesions exhibit region-specific characteristics that determine their clinical significance~\cite{brugge2015diagnosis}. Current whole-pancreas image analysis approaches fail to capture these critical regional variations, substantially limiting the effectiveness of machine learning (ML) classification for diseases that selectively target specific pancreatic regions rather than the entire organ.

Despite its clear clinical importance, pancreas part segmentation—dividing the organ into \textit{head, body, and tail}—represents an unaddressed technical challenge in medical image analysis~\cite{zhang2023deep}. The task is exceptionally complex due to the variable morphology of pancreatic parts, with the head aligned craniocaudally while the body and tail orient horizontally. This anatomical variability creates intricate spatial relationships that make part segmentation substantially more challenging than conventional whole-organ approaches~\cite{tirkes2012mr}. The technical novelty of our work lies in developing specialized deep learning (DL) architectures that can effectively model these complex inter-part relationships while maintaining anatomical consistency. 

A fundamental obstacle to advancing this field has been the complete absence of well-curated, manually segmented datasets for pancreatic part segmentation in the literature. While the literature in DL based part segmentation is completely missing, we also recognize the first work in this topic with registration and atlas-based approaches. This data scarcity, compounded by stringent privacy regulations and institutional barriers to data sharing, has created a significant impediment to training robust DL models~\cite{shokri2015privacy}. FL has emerged as a promising solution to these challenges, enabling collaborative model training across multiple institutions without the need for direct data sharing~\cite{li2020federated}. In the literature, FL has already demonstrated value in several pancreatic imaging projects, including pancreatic cyst risk classification, while maintaining patient data privacy and regulatory compliance~\cite{pan2024adaptive,pan2024ipmn}. Our innovative FL approach directly addresses this challenge, enabling collaborative model training across seven diverse institutions without compromising patient privacy.

In this study, we explore the application of FL for pancreas part segmentation using a large multi-center MRI dataset comprising 711 T1W and 726 T2W images. We systematically evaluate three state-of-the-art segmentation architectures with two specialized FL algorithms, quantifying performance through comprehensive segmentation metrics. Our key contributions include:

\begin{itemize}
    \item First-ever implementation of anatomically-informed pancreas part segmentation (head, body, tail) achieving clinically viable accuracy, enabling precise characterization of region-specific pancreatic pathologies.
    \item Novel application of specialized DL architectures to address the unique challenges of MRI-based pancreas part segmentation, overcoming limitations of conventional whole-organ CT approaches.
    \item FL framework that maintains data privacy while preserving segmentation performance across heterogeneous, multi-institutional datasets.
\end{itemize}

This study bridges two under-explored gaps: (1) FL’s utility for complex multi-region pancreas segmentation in MRI, and (2) the interplay between anatomical priors and federated optimization. By enabling privacy-preserving collaboration across institutions, our framework accelerates the development of robust models for regionally-manifesting pancreatic diseases, with immediate applicability to pancreatic disease diagnostics.

\begin{figure}[htbp]
\includegraphics[width=\textwidth]{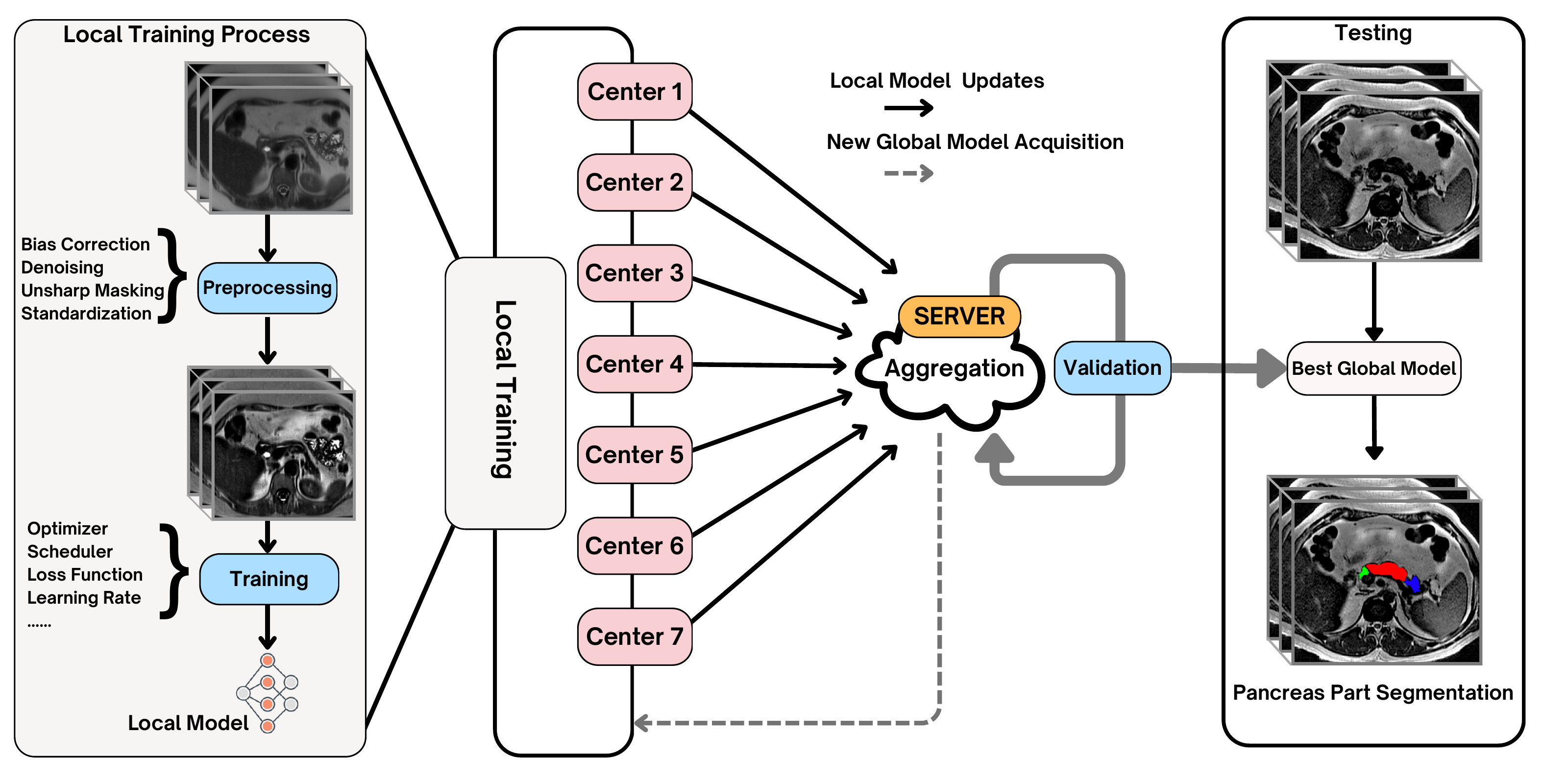}
\caption{This workflow illustrates pancreas part segmentation using FL. Each institution independently preprocesses its data and trains a local model using the same optimization strategy. After each training epoch, local models are sent to a central server for aggregation, forming a new global model. The global model is then validated, and the best-performing version is selected based on the Dice score. Finally, the best model is used for pancreas part segmentation on test images.} \label{workflow}
\end{figure}

\section{Methodology}
\subsection{Federated Learning}
\subsubsection{FedAvg}
Federated Averaging (FedAvg) is the mostly used optimization algorithm in FL hat enables collaborative model training without exposing sensitive raw data~\cite{mcmahan2017communication}. The FedAvg with \textit{K} centers can be described as the local updated step and Global aggregation step. The FedAvg algorithm operates through an iterative two-phase process: distributed local training followed by global parameter aggregation. During each training epoch, the algorithm maintains model architectural consistency while allowing for computational and statistical heterogeneity across participating sites. Specifically, for our multi-institutional pancreas MRI dataset distributed across \textit{K} medical centers, the algorithm proceeds as follows: 

\textbf{Local Updated}: At each client (e.g., hospital $k$), the local model minimizes a region-aware Dice loss tailored for pancreas parts:
\begin{equation}
    F_k(w) = \frac{1}{N_k} \sum_{i \in D_k} \sum_{r}\ell_{r}\bigl(f(x_i; w),\, y_{i,r}\bigr),
\end{equation}
where $r$ is the pancreas region (head, body, tail), $y_{i,r}$ is ground truth mask for  pancreas region $r$, $N_k$ is the quantity of data samples in center $k$, $D_k$ represents the dataset for center $k$, and $\ell_{r}(\cdot)$ is the Dice based loss function corresponding to the pancreas region $r$. Unlike standard FedAvg, our loss explicitly weights regions by their clinical salience (e.g., head tumors dominate prognosis), countering MRI’s low contrast at part boundaries. Furthermore, $f(x_i; w)$ computes the model output for input $x_i$ with parameters $w$.

\textbf{Global Aggregation}:
The aggregation of the local models follows a weighted averaging approach, which can be written as:
\begin{equation}
\mathbf{w_{global}} = \frac{\sum_{k=1}^{K} N_k \mathbf{w}_k}{\sum_{k=1}^{K} N_k},
\end{equation}
where the $w$ is the weight updated from center $k$ and $N_K$ accounts for client data imbalance (e.g., large hospitals vs. rural clinics). While effective under IID data, this approach struggles with directional inconsistency in non-IID settings: gradients from clients with tail-dominant pathologies (e.g., chronic pancreatitis) conflict with those from head-tumor cohorts. 

\subsubsection{FedProx}
FedAvg’s assumption of IID data might be violated in our task, as pancreatic morphology and pathology distributions vary significantly across institutions (e.g., head tumors are prevalent in Site A vs. tail cysts in Site B). This leads to model degradation—empirically. To mitigate this, we benchmark FedAvg against FedProx~\cite{li2020federated}, which adds a proximal term to align local and global objectives:
\begin{equation}
    F_{k}^{\text{prox}}(w) = F_{k}(w) + \frac{\mu}{2} \,\|w - w_{global}\|^2,
\end{equation}
where $w_t$ is the global model from the previous communication round, 
$\mu \ge 0$ is a tuning parameter that controls the regularization strength, 
and $\|\cdot\|$ denotes the $L^2$-norm. The proximal term $\tfrac{\mu}{2}\,\|w - w_t\|^2$ penalizes large deviations from the global model, improving stability in non-iid settings. 
This modification helps mitigate variability due to heterogeneity in data distributions among institutions.

\subsection{Data Collection and Pre-processing}
We used publicly available multi-center pancreatic MRI dataset collected from seven institutions, consisting of 711 T1W and 726 T2W images~\cite{zhang2025large}. Two abdominal radiologists from our team manually delineated each pancreas in this dataset into the pancreatic head, body, and tail parts using a standardized protocol established in 3D Slicer (version 5.6.2)~\cite{fedorov20123d}. This manual segmentation process was time-consuming and labor-intensive, requiring approximately 40 minutes per case to delineate pancreatic regions. All segmentation masks underwent review by a senior abdominal radiologist to ensure quality and accuracy. The pancreatic part boundaries were determined based on both established anatomical landmark visible in cross-sectional imaging and consensus definitions developed by our team of abdominal radiologists:
\begin{itemize}
    \item The head-body boundary was marked at the medial border of the superior mesenteric vein~\cite{desouza2018quantitative},
    \item The body-tail boundary was defined by splitting the remaining pancreatic tissue in half on longitudinal measurement, as there is no universally determined boundary in the literature.
\end{itemize}
Images with severe anatomical disruption, massive tissue loss, or other significant abnormalities that prevented reliable part segmentation were discussed in consensus and excluded from the dataset to maintain annotation consistency and quality.

To enhance the quality and consistency of the input data, we applied a three-step pre-processing pipeline consisting of Gaussian denoising, unsharp masking, and intensity standardization~\cite{nyul1999standardizing}. The last step (intensity standardization) mitigates variations due to acquisition differences and improves generalizability across datasets.  

\subsection{Pancreas Part Segmentation}
We evaluate three advanced deep learning architectures for pancreas part segmentation within our FL framework: the classical U-Net with its symmetric encoder-decoder structure and skip connections for preserving spatial information~\cite{ronneberger2015u}; Attention U-Net, which integrates attention gates to dynamically highlight salient features critical for delineating the subtle boundaries between pancreatic regions~\cite{oktay2018attention}; and Swin-UNETR, a recent innovation leveraging hierarchical Swin Transformer blocks to capture the complex spatial relationships and variable morphology unique to pancreatic anatomy~\cite{hatamizadeh2021swin}. Our technical evaluation employs a rigorous multi-metric assessment framework to quantify segmentation performance across the challenging boundaries of pancreatic head, body, and tail regions. The metrics include Dice coefficient to measure volumetric overlap, average symmetric surface distance (ASSD) and 95th percentile Hausdorff distance (HD95) to evaluate boundary delineation accuracy, and precision-recall analysis to assess regional classification performance. This comprehensive evaluation is essential given the subtle tissue contrast variations and complex morphological boundaries characteristic of pancreatic regions in MRI.

%\subsection{Implementation Details}
\textbf{Implementation Details.} Fig~\ref{workflow} illustrates our FL experiment workflow. We randomly selected 10\%  of the data from each modality for the test. The remaining data set was divided into 80\% for training and 20\% for validation to ensure a balanced evaluation. To rigidly assess their performance, all models are trained from scratch and present a convergent loss at the end of training. 

For centralized learning, we utilized the Medical Open Network for Artificial Intelligence(MONAI) framework from NVIDIA\cite{cardoso2022monai}. The models were trained for 400 epochs using the Dice loss function and the AdamW optimizer~\cite{loshchilov2017decoupled}. To ensure optimal model performance, we saved the model with the best average Dice score across training. For FL, we implemented FedAvg and FedProx as aggregation strategies while maintaining the same training approach used in centralized learning. We applied both FedAvg and FedProx methods to three different representative segmentation architectures: U-Net~\cite{ronneberger2015u}, Attention U-Net~\cite{oktay2018attention}, and Swin-UNETR~\cite{hatamizadeh2021swin}, ensuring a comprehensive evaluation of federated learning in medical image segmentation.

\section{Experiment Results}
Table~\ref{tab:average} presents the average segmentation performance across the pancreas head, body, and tail for both T1-weighted and T2-weighted images under centralized and federated learning settings.

For centralized models, Attention U-Net demonstrated superior performance on T1-weighted images with a Dice score of 0.65, leveraging its attention mechanism to effectively capture the subtle boundaries between pancreatic regions. In contrast, Swin-UNETR excelled on T2-weighted images (Dice: 0.78), where its transformer-based architecture effectively captured the long-range dependencies critical for distinguishing the complex morphological transitions between pancreatic head, body, and tail. Quantitatively, Swin-UNETR outperformed competing architectures in boundary delineation metrics (ASSD, HD95) and region coverage (IoU, recall), while Attention U-Net exhibited marginally higher precision.

A key finding of our study is the differential impact of FL across imaging sequences. For T1-weighted images, our FL approach notably enhanced segmentation performance, with Attention U-Net+FedAvg achieving the highest Dice score (0.67), followed closely by U-Net+FedAvg (0.66) and Swin-UNETR+FedAvg (0.65). This performance improvement contradicts conventional expectations, as federated learning typically underperforms centralized training due to limited data access.

\begin{table}[]
    \centering
    \setlength{\tabcolsep}{1.5pt} %
    \renewcommand{\arraystretch}{1} % 
    \caption{Average segmentation performance across all pancreas head, body, and tail. ASSD and HD95 are shown in millimeters.}
    \label{tab:average}
    % \begin{tabular}{l p{1.2cm} p{1.2cm} p{1.2cm} p{1.2cm} p{1.4cm} p{1.4cm}}
    \scalebox{0.85}{
    \begin{tabular}{l |cccccc|cccccc}
    \bottomrule

     \rowcolor{g1}& \multicolumn{6}{c|}{\textbf{T1 Weighted}} & \multicolumn{6}{c}{\textbf{T2 Weighted}} \\
    \rowcolor{g1} \multirow{-2}{*}{\textbf{Method}} & Dice & ASSD & HD95 & mIoU & Precision & Recall & Dice & ASSD & HD95 & mIoU & Precision & Recall \\\hline
       
        % \toprule
       
        U-Net               &$0.62$  & $11.02$ & $3.60 $ & $0.49 $ & $0.79 $ & $0.55 $  &$0.70 $ & $7.00 $ & $1.26 $ & $0.57 $ & $0.83 $ & $0.63 $              \\
        +FedAvg             &$0.66$  & $8.95$ & $2.48$ &  $\textbf{0.53}$ & $0.79$&  $0.60$       &$0.70 $ & $6.95 $ & $1.08 $ & $0.55 $ & $0.82 $ & $0.63 $                \\
        +FedProx           &$0.66 $ & $8.49 $ & $\textbf{2.03} $ & $\textbf{0.53} $ & $0.78 $ & $\textbf{0.61} $  &$0.73 $ & $5.38 $ & $0.72 $ & $0.59 $ & $0.85 $ & $0.66 $      \\
        \hline
        Swin-UNETR              &$0.63 $ & $10.78 $ & $3.72 $ & $0.49 $ & $0.79 $ & $0.56 $        & $\textbf{0.78} $ &  $\textbf{4.11 }$ &  $\textbf{0.57 }$ &  $\textbf{0.65} $ &  $\textbf{0.89 }$ &  $\textbf{0.71 }$        \\
        +FedAvg       &$0.65 $ & $9.34$ & $2.26 $ & $0.51 $ & $0.80 $ & $0.58 $       &$0.72 $ & $7.79 $ & $2.14 $ & $0.58 $ & $0.84 $ & $0.66 $          \\
        +FedProx   &$0.61 $ & $11.76 $ & $4.22 $ & $0.47 $ & $0.76 $ & $0.54 $   &$0.62 $ & $10.08 $ & $2.23 $ & $0.47 $ & $0.75 $ & $0.56 $      \\
        \hline
        Attention U-Net         &$0.65 $ & $9.35 $ & $3.26 $ & $0.51 $ & $0.80 $ & $0.57 $     & $\textbf{0.78} $ & $4.26 $ & $0.63 $ &  $\textbf{0.65} $ & $0.88 $ &  $\textbf{0.71} $           \\
        +FedAvg  & $\textbf{0.67}$ &  $\textbf{8.30} $ & $2.15  $ &  $\textbf{ 0.53} $ &  $\textbf{0.81}  $ &  $0.60 $       &$0.74 $ & $5.30 $ & $0.81 $ & $0.60 $ & $0.85 $ & $0.67 $  \\
        +FedProx &$0.66 $ & $9.82 $ & $3.57 $ & $0.52 $ & $0.79 $ & $0.59 $  &$0.70 $ & $6.51 $ & $0.90 $ & $0.56 $ & $0.83 $ & $0.64 $   \\
        \hline
    \end{tabular}
    }
    \vspace{10pt}
    \centering
    \setlength{\tabcolsep}{1.5pt} %
    \renewcommand{\arraystretch}{1} % 
    \caption{Segmentation performance on pancreas head, body, and tail.}
    \label{tab:segmentation of parts}
    % \begin{tabular}{l p{1.2cm} p{1.2cm} p{1.2cm} p{1.2cm} p{1.4cm} p{1.4cm}}
    \scalebox{0.85}{
    \begin{tabular}{l |cccccc|cccccc}
        \bottomrule
        
         \rowcolor{g1}& \multicolumn{6}{c|}{\textbf{T1 Weighted}} & \multicolumn{6}{c}{\textbf{T2 Weighted}} \\
        \rowcolor{g1}\multirow{-2}{*}{\textbf{Method}}& Dice & ASSD & HD95 & mIoU & Precision & Recall & Dice & ASSD & HD95 & mIoU & Precision & Recall \\\hline
        % \bottomrule
        % \toprule

        \rowcolor{g3}\multicolumn{13}{c}{\textbf{Pancreas Head}} \\\hline
        % \midrule
        U-Net               &$0.68 $ & $9.41 $ & $2.94 $ & $0.55 $ & $0.85 $ & $0.60 $ &$0.78 $ & $4.77 $ & $0.74 $ & $0.65 $ & $0.89 $ & $0.71 $               \\
        +FedAvg             &$0.72 $ & $9.81 $ & $4.50 $ & $0.58 $ & $0.84 $ & $0.64 $&$0.77 $ & $5.00 $ & $0.75 $ & $0.64 $ & $0.88 $ & $0.71 $                \\
        +FedProx            &$\textbf{0.73} $ & $6.46 $ & $1.70 $ & $\textbf{0.60} $ & $0.80 $ & $\textbf{0.69} $ &$0.76 $ & $5.16 $ & $0.80 $ & $0.62 $ & $0.87 $ & $0.69 $       \\
        \hline
        Swin-UNETR    &$0.64 $ & $7.85 $ & $1.76 $ & $0.51 $ & $0.84 $ & $0.56 $&$0.79 $ &  $\textbf{4.09} $ & $0.69 $ & $0.67 $ &  $\textbf{0.92} $ & $0.72 $        \\
        +FedAvg       &$0.72 $ & $6.41 $ &  $\textbf{1.01} $ & $0.58 $ & $0.86 $ &  $0.64 $&$0.75 $ & $5.66 $ & $0.85 $ & $0.62 $ & $0.87 $ & $0.69 $         \\
        +FedProx      &$0.66 $ & $9.38 $ & $3.18 $ & $0.52 $ & $0.81 $ & $0.58 $   &$0.67 $ & $7.20 $ & $1.23 $ & $0.52 $ & $0.80 $ & $0.62 $      \\
        \hline
        Attention U-Net         &$0.67 $ & $9.31 $ & $3.14 $ & $0.53 $ & $0.86 $ & $0.57 $& $\textbf{0.80} $ &  $\textbf{4.09} $ &  $\textbf{0.63} $ &  $\textbf{0.68} $ & $0.91 $ &  $\textbf{0.73} $           \\
        +FedAvg                 & $\textbf{0.73} $ &  $\textbf{6.23} $ &  $\textbf{1.01} $ &  $0.59 $ &  $\textbf{0.88} $ &  $0.64 $ &$0.75 $ & $5.88 $ & $1.15 $ & $0.62 $ & $0.89 $ & $0.67 $  \\
        +FedProx &$0.71 $ & $8.48 $ & $3.01 $ & $0.56 $ & $0.85 $ & $0.62 $   &$0.74 $ & $5.61 $ & $0.88 $ & $0.61 $ & $0.86 $ & $0.68 $   \\
        
        \hline
        \rowcolor{g3}\multicolumn{13}{c}{\textbf{Pancreas Body}} \\
        \hline
        U-Net               &$0.62 $ & $9.74 $ & $1.72 $ & $0.47 $ & $\textbf{0.81} $ & $0.53 $&$0.70 $ & $7.05 $ & $0.88 $ & $0.56 $ & $0.82 $ & $ 0.14$ \\
        +FedAvg             &$0.65 $ & $8.91 $ & $1.78 $ & $0.51 $ & $0.78 $ & $0.58 $&$0.69 $ & $7.01 $ & $0.82 $ & $0.54 $ & $0.82 $ & $0.62 $ \\
        +FedProx            &$0.64 $ & $8.53 $ & $1.37 $ & $0.50 $ & $0.78 $ & $0.57 $   &$0.72 $ & $5.59 $ & $0.60 $ & $0.58 $ & $0.85 $ & $0.65 $     \\
        \hline
        Swin-UNETR    &$0.60 $ & $12.75 $ & $4.08 $ & $0.46 $ & $0.78 $ & $0.53 $ &$\textbf{0.77} $ & $4.41 $ & $\textbf{0.48} $ & $\textbf{0.64} $ & $0.89 $ & $\textbf{0.70} $ \\
        +FedAvg      &$0.63 $ & $9.14 $ & $\textbf{1.21} $ & $0.48 $ & $0.80 $ & $0.55 $ &$0.71 $ & $9.80 $ & $3.69 $ & $0.57 $ & $0.84 $ & $0.65 $\\
        +FedProx    &$0.60 $ & $12.30 $ & $4.03 $ & $0.46 $ & $0.74 $ & $0.53 $ &$0.61 $ & $10.21 $ & $1.08 $ & $0.47 $ & $0.77 $ & $0.54 $      \\
        \hline
        Attention U-Net         &$0.64 $ & $8.20 $ & $1.71 $ & $0.49 $ & $\textbf{0.81} $ & $0.56 $&$\textbf{0.77} $ & $\textbf{4.33} $ & $\textbf{0.48} $ & $\textbf{0.64} $ & $\textbf{0.90} $ & $0.68 $           \\
        +FedAvg                 &$\textbf{0.66} $ & $\textbf{7.35} $ & $1.37 $ & $\textbf{0.52} $ & $0.80 $ & $\textbf{0.59} $ &$0.75 $ & $5.03 $ & $0.55 $ & $0.61 $ & $0.86 $ & $0.68 $ \\ 
        +FedProx  &$0.65 $ & $8.35 $ & $1.37 $ & $0.50 $ & $0.78 $ & $\textbf{0.59} $   &$0.71 $ & $6.04 $ & $0.68 $ & $0.56 $ & $0.84 $ & $0.65 $   \\
        \hline
        \rowcolor{g3}\multicolumn{13}{c}{\textbf{Pancreas Tail}} \\
        \hline
        U-Net               &$0.58 $ & $13.91 $ & $6.13 $ & $0.45 $ & $0.72 $ & $0.52 $&$0.63 $ & $9.18 $ & $2.15 $ & $0.49 $ & $0.78 $ & $0.55 $ \\
        +FedAvg             &$0.61 $ & $11.31 $ & $3.98 $ & $0.48 $ & $0.74 $ & $0.55 $ &$0.63 $ & $8.84 $ & $1.67 $ & $0.48 $ & $0.77 $ & $0.56 $\\
        +FedProx            &$0.61 $ & $10.71 $ & $\textbf{2.98} $ & $0.47 $ & $0.73 $ & $0.55 $    &$0.71 $ & $5.40 $ & $0.75 $ & $0.56 $ & $0.82 $ & $0.64 $    \\
        \hline
        Swin-UNETR     &$0.61 $ & $12.71 $ & $5.42 $ & $0.48 $ & $0.72 $ & $0.56 $&$\textbf{0.78} $ & $\textbf{3.83} $ & $\textbf{0.54} $ & $\textbf{0.65} $ & $\textbf{0.86} $ & $\textbf{0.72} $\\
        +FedAvg      &$0.60 $ & $12.64 $ & $4.54 $ & $0.47 $ & $0.75 $ & $0.54 $ &$0.69 $ & $7.92 $ & $1.87 $ & $0.55 $ & $0.80 $ & $0.63 $\\
        +FedProx    &$0.55 $ & $15.41 $ & $7.01 $ & $0.43 $ & $0.69 $ & $0.49 $ &$0.56 $ & $12.81 $ & $4.38 $ & $0.42 $ & $0.68 $ & $0.52 $        \\
        \hline
        Attention U-Net    &$\textbf{0.62} $ & $\textbf{10.63} $ & $4.77 $ & $\textbf{0.49} $ & $0.73 $ & $0.56 $   &$0.76 $ & $4.38 $ & $0.77 $ & $0.63 $ & $0.82 $ & $\textbf{0.72} $       \\
        +FedAvg            &$\textbf{0.62} $ & $11.27 $ & $4.05 $ & $\textbf{0.49} $ & $\textbf{0.76} $ & $\textbf{0.57} $   &$0.72 $ & $4.99 $ & $0.73 $ & $0.58 $ & $0.80 $ & $0.67 $   \\
        +FedProx &$0.60 $ & $12.86 $ & $6.26 $ & $0.47 $ & $0.73 $ & $0.55 $  &$0.65 $ & $7.88 $ & $1.14 $ & $0.50 $ & $0.78 $ & $0.58 $     \\
        \hline
    \end{tabular}
    }
\end{table}

\begin{figure}[htbp]
\includegraphics[width=\textwidth]{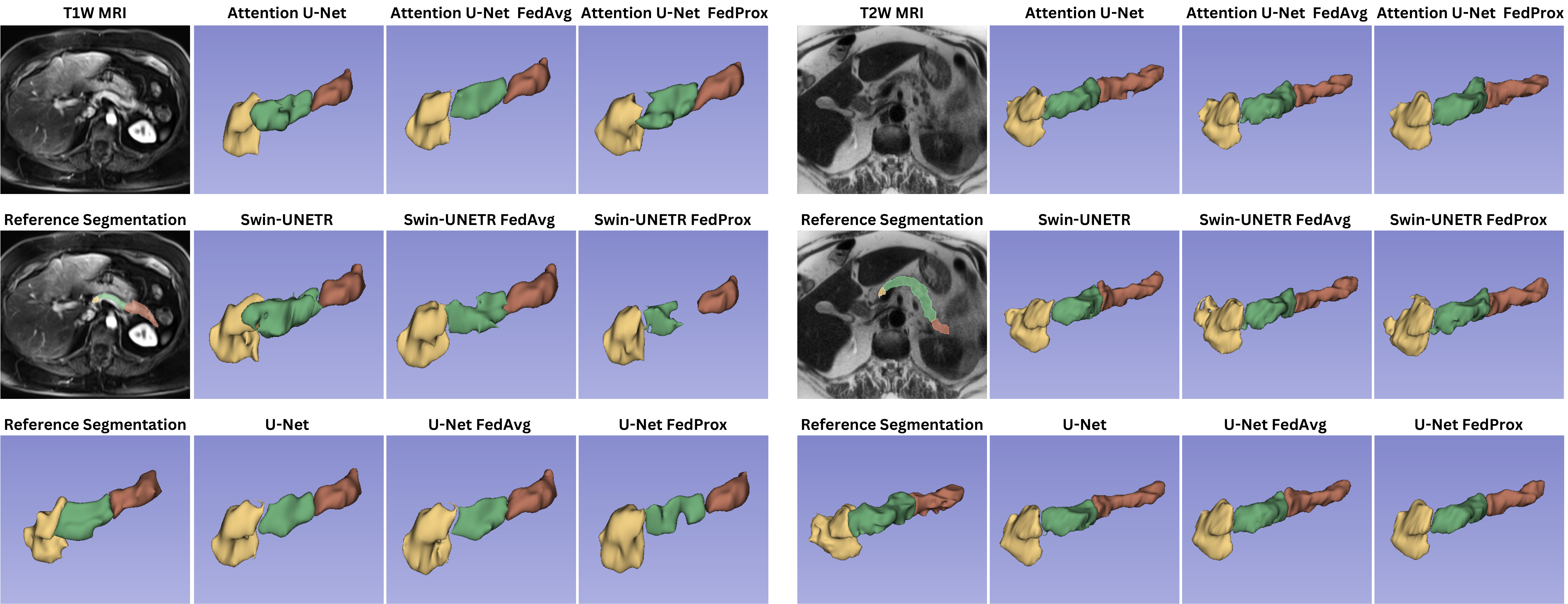}
\caption{Pancreas part segmentation samples: The left side displays the T1W image along with the corresponding segmentation results from each model, while the right side presents the T2W image and its corresponding segmentation results.} \label{MRI}
\end{figure}

For T2-weighted images, we observed a more nuanced pattern: U-Net maintained consistent performance in federated settings, while transformer-based architectures (Swin-UNETR, Attention U-Net) showed modest performance decreases. Interestingly, boundary metrics (HD95, ASSD) improved under federated learning, suggesting enhanced generalization at region interfaces. This finding indicates that federated learning's inherently regularized optimization trajectory may mitigate overfitting in certain contexts, particularly for T1W images where tissue contrast is inherently lower.

Table~\ref{tab:segmentation of parts} presents a region-specific analysis that reveals differential performance across the three anatomical subdivisions. Federated learning consistently enhanced metrics across all pancreatic regions compared to standalone models. Advanced architectures (Swin-UNETR, Attention U-Net) demonstrated superior performance in capturing region-specific features, particularly in recall and precision metrics that are critical for clinical applications.

Fig~\ref{MRI} provides visual confirmation of these quantitative findings, with color-coded segmentations (head: yellow, body: green, tail: red) across both T1W and T2W MRI images. The visualizations reveal a performance gradient across regions, with the pancreas head achieving the highest accuracy while the tail exhibits greater variability. This performance differential aligns with the anatomical challenges inherent in pancreatic part segmentation, where the tail region's variable morphology and less distinct boundaries present greater segmentation challenges. These findings emphasize the need for region-adaptive segmentation strategies in future algorithm development.

\section{Conclusion}
In this work, we employed two federated learning algorithms with three state-of-the-art models to preserve data privacy. We comprehensively evaluated the performance of the U-Net, Swin-UNETR, and Attention U-Net on the pancreas part segmentation and this is the first implementation. The experimental results show that federated learning can achieve similar performance compared to centralized learning, demonstrating the potential of employing federated algorithms in practical applications. This also provides valuable insight into future collaborations between centers.

%\subsubsection{Acknowledgments}

% \bibliography{ref} 

% \end{thebibliography}
\end{document}